\documentclass[runningheads]{llncs}
\usepackage{graphicx}
\usepackage{tikz}
\usepackage{comment}
\usepackage{amsmath,amssymb} % define this before the line numbering.
\usepackage{color}
\usepackage{booktabs}
\usepackage{orcidlink}

\usepackage{graphicx}
\usepackage{amsmath}
\usepackage{amssymb}
\usepackage{booktabs}

\usepackage{makecell}
\usepackage{colortbl}
\usepackage{xcolor}
\usepackage{multirow}
\usepackage{bbding}
\usepackage{pifont}
\usepackage{wasysym}
\usepackage{todonotes}
\usepackage[normalem]{ulem}
\usepackage{lipsum}

\usepackage[font=small]{caption} 
\makeatletter
\newcommand{\printfnsymbol}[1]{%
  \textsuperscript{\@fnsymbol{#1}}%
}

\newcommand\freefootnote[1]{%
  \let\thefootnote\relax%
  \footnotetext{#1}%
  \let\thefootnote\svthefootnote%
}

\begin{document}
% \renewcommand\thelinenumber{\color[rgb]{0.2,0.5,0.8}\normalfont\sffamily\scriptsize\arabic{linenumber}\color[rgb]{0,0,0}}
% \renewcommand\makeLineNumber {\hss\thelinenumber\ \hspace{6mm} \rlap{\hskip\textwidth\ \hspace{6.5mm}\thelinenumber}}
% \linenumbers
% \pagestyle{headings}
% \mainmatter
% \def\ECCVSubNumber{7351}  % Insert your submission number here

%\title{Toward Realistic sport video Captioning: An Identity-Aware Large-Scale Basketball Datasets with Cross-Feature Transformer} % Replace with your title
% \title{Sport Video Analysis on Large-Scale Data}
% INITIAL SUBMISSION 
%\begin{comment}
% \titlerunning{ECCV-22 submission ID \ECCVSubNumber} 
% \authorrunning{ECCV-22 submission ID \ECCVSubNumber} 
% \author{Anonymous ECCV submission}
% \institute{Paper ID \ECCVSubNumber}
%\end{comment}
%******************

% CAMERA READY SUBMISSION
% \begin{comment}
% \authorrunning{}
\authorrunning{D. Wu, H. Zhao, et al.}
\title{Sports Video Analysis on Large-Scale Data}
\titlerunning{Sports Video Analysis on Large-Scale Data}
% If the paper title is too long for the running head, you can set
% an abbreviated paper title here
%
% \author{Dekun Wu^{*}\inst{1}\orcidID{0000-1111-2222-3333} \and
% He Zhao^{*}\inst{2}\orcidID{1111-2222-3333-4444} \and
% Xingce Bao\inst{3}\orcidID{2222--3333-4444-5555}
% \and Richard P. Wildes\inst{2}}
% \author{Dekun Wu\thanks{equal contribution}
% \and He Zhao\printfnsymbol{1} \and Xingce Bao \and Richard P. Wildes \and \\
% University of Pittsburgh \and York University\\ \and
% \'Ecole Polytechnique F\'ed\'erale de Lausanne (EPFL)\\ \and
% {\tt\small dew104@pitt.edu},
% {\tt\small \{zhufl, wildes\}@cse.yorku.ca},
% {\tt\small xingce.bao@alumni.epfl.ch}
% }

\begin{comment}
he zhao orcid: 0000-0002-2471-9680,
dekun wu orcid: 0000-0002-5518-093X,
xingce bao orcid: 0000-0001-7458-8052,
rich orcid: 0000-0003-3433-1329,
\end{comment}

\author{Dekun Wu\thanks{Equal contribution}$\mathbf{^{1}}$\orcidlink{0000-0002-5518-093X}
\and
He Zhao\printfnsymbol{1}$\mathbf{^{2}}$\orcidlink{0000-0002-2471-9680}
\and
Xingce Bao$\mathbf{^{3}}$\orcidlink{0000-0001-7458-8052}
\and
Richard P. Wildes$\mathbf{^{2}}$\orcidlink{0000-0003-3433-1329}
% \vspace{6pt}
\\
{ $\mathbf{^1}$University of Pittsburgh, $\mathbf{^2}$York University
\\
 $\mathbf{^3}$\'Ecole Polytechnique F\'ed\'erale de Lausanne (EPFL)}
% }
% \vspace{3pt}
\\
%First line of institution2 address\\
{\tt\small dew104@pitt.edu},
{\tt\small \{zhufl, wildes\}@cse.yorku.ca},\\
{\tt\small xingce.bao@alumni.epfl.ch}
% {\tt\small \{zhufl, derpanis, wildes\}@cse.yorku.ca},\\
% {\tt\small \{isma.hadji, n.dvornik, allan.jepson\}@samsung.com}
}
\institute{}
% \authorrunning{}

% \maketitle

% \institute{University of Pittsburgh \and
% York University\\ \and
% \'Ecole Polytechnique F\'ed\'erale de Lausanne (EPFL)}
% \end{comment}
%******************
\maketitle

\begin{abstract}
This paper investigates the  modeling of automated machine description on sports video, which has seen much progress recently. Nevertheless, state-of-the-art approaches fall quite short of capturing how human experts analyze sports scenes. There are several major reasons: (1) The used dataset is collected from non-official providers, which naturally creates a gap between models trained on those datasets and real-world applications; (2) previously proposed methods require extensive annotation efforts (i.e., player and ball segmentation at pixel level) on localizing useful visual features to yield acceptable results; (3) very few public datasets are available. In this paper, we propose a novel large-scale NBA dataset for Sports Video Analysis (NSVA) with a focus on captioning, to address the above challenges. We also design a unified approach to process raw videos into a stack of meaningful features with minimum labelling efforts, showing that cross modeling on such features using a transformer architecture leads to strong performance. In addition, we demonstrate the broad application of NSVA by addressing two additional tasks, namely fine-grained sports action recognition and salient player identification.
Code and dataset are available at \url{https://github.com/jackwu502/NSVA}.
\end{abstract}

\freefootnote{Corresponding author: dew104@pitt.edu}

\section{Introduction}
\begin{figure}[t]
\centering
\includegraphics[width=1.0\linewidth]{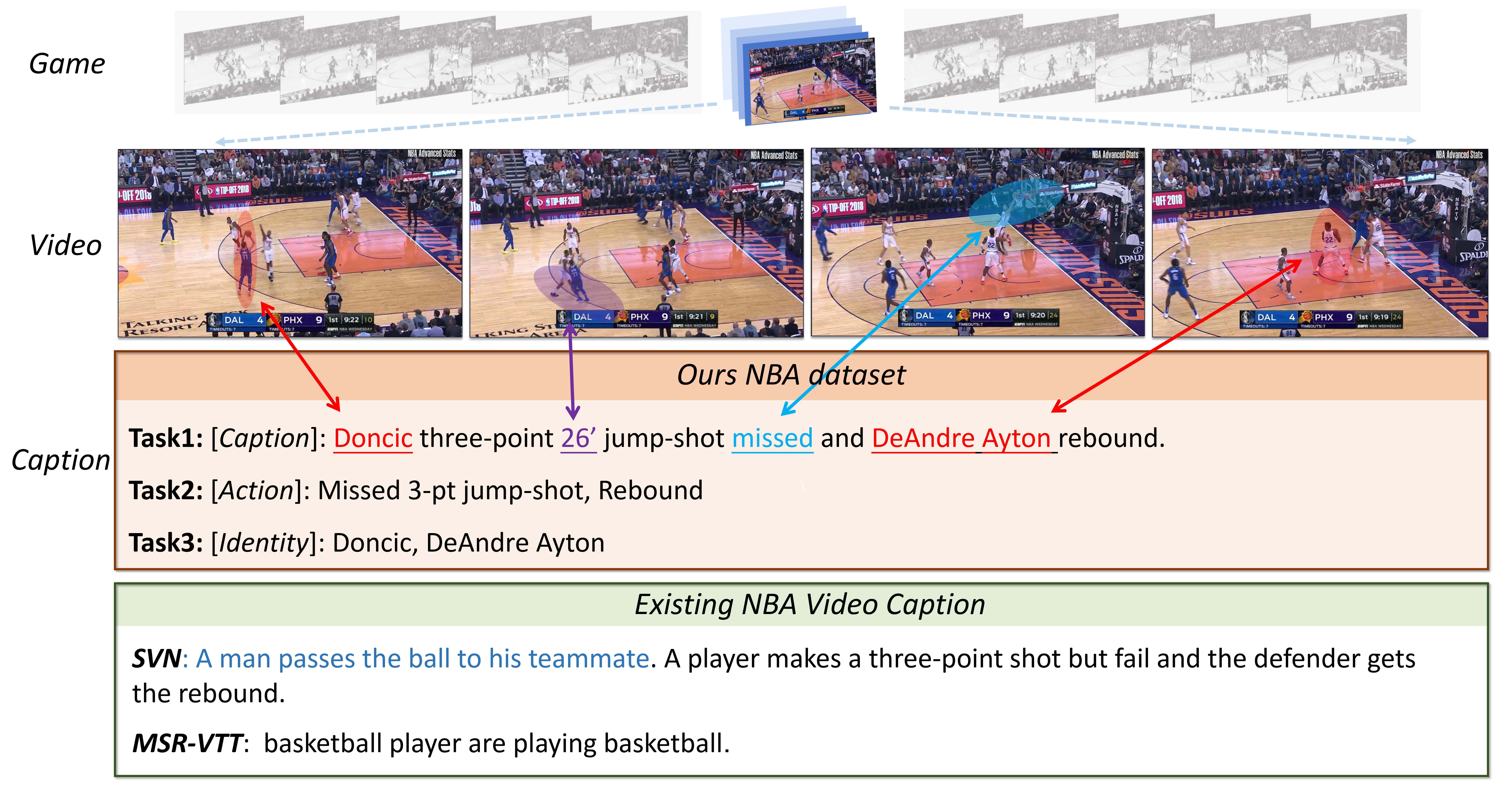}
\caption{
Which one is more descriptive for the above professional sport game clip?
Conceptual comparison between NSVA (top box)
and
extant basketball (NBA) video captioning datasets \cite{Yu-cvpr2018,msrvtt} (bottom box). The sentence in blue text describes a passing action, which might not be practically valuable and is not a focus of NSVA. Instead, captions in NSVA target compact information that could enable statistics counting and game analysis.
Moreover, both alternative captioning approaches lack in important detail (e.g., player identities and locations).
}
\label{fig:teaser}
\end{figure}

Recently, there have been many attempts aimed at empowering machines to describe the content presented in a given video~\cite{krishna-iccv2017,DBLP:journals/corr/abs-1806-08854,Yu-cvpr2018,svcdv}. The particular challenge of generating a text from a given video is termed ``video captioning''~\cite{Aafaq-Computing-Surveys-2019}. Sports video captioning is one of the most intriguing video captioning sub-domains, as sports videos usually contain multiple events depicting the interactions between players and objects, e.g., ball, hoop and net. Over recent years, many efforts have addressed the challenge of sports video captioning for soccer, basketball and volleyball games~\cite{svcdv,Yu-cvpr2018,SVN}.

Despite the recent progress seen in sports video captioning, previous efforts share three major limitations. (1) They all require laborious human annotation efforts that limit the scale of data~\cite{svcdv,Yu-cvpr2018,SVN}. (2) Some previous efforts do not release data~\cite{svcdv,Yu-cvpr2018,SVN}, and thereby prevent others from accessing useful data resources. (3) The collected human annotations typically lack the diversity of natural language and related intricacies. Instead, they tend to focus on details that are not interesting to human viewers, e.g. passing or dribbling activities (see Figure~\ref{fig:teaser}), while lacking important information (e.g. identity of performing players). In this regard, a large-scale sports video dataset that is readily accessible to researchers and annotated by professional sport analysts is very much needed. In response we propose NBA dataset for Sports Video Analysis (NSVA). 
%and we introduce NSVA (NBA dataset for sport video analysis).
%To thi s end, we introduce NSVA, a NBA dataset for sport video analysis.

Figure \ref{fig:teaser} shows captions depicting the same sports scene from NSVA, MSR-VTT \cite{msrvtt} and another fine-grained sports video captioning dataset, SVN \cite{Yu-cvpr2018}. Our caption is compact, focuses on key actions (e.g., \textit{made shot}, \textit{miss shot} and \textit{rebound}) and is identity aware. Consequently, it could be further translated to a box score for keeping player and team statistics. SVN includes more less important actions, e.g., \textit{passing}, \textit{dribbling} or \textit{standing}, which are excessively common but of questionable necessity. They neither cover player names nor essential details, e.g., shooting from 26 feet away. This characteristic of NSVA poses a great challenge as it requires models to ignore spatiotemporally dominant, yet unimportant, events and instead focus on key events that are of interest to viewers, even though they might have unremarkable visual presence. Additionally, NSVA also requires the model to identify the players whose actions will be recorded in the box score. This characteristic adds another difficulty to NSVA and distinguishes us from all previous work, where player identification is under-emphasized by only referring to ``a man'', ``some player'', ``offender'', etc.
%In this paper, we show how we solve the problem by devising features specifically for this task and proposing several multi-feature fusion strategies to combining these features for the sake of performance. Extensive experiments demonstrate the effectiveness of our proposed method in this challenging task and the superiority of our method to the state of the arts. 

\textbf{Contributions.} The contributions of this paper are threefold. (1) We propose a new identity-aware NBA dataset for sports video analysis (NSVA), which is built on web data, to fill the vacancy left by previous work whose datasets are neither identity aware nor publicly available. (2) Multiple novel features are devised, especially for modeling captioning as supported by NSVA, and are used for input to a unified transformer framework. Our designed features can be had with minimal annotation expense and provide complementary kinds of information for sports video analysis. 
%Our proposed method has been proven very effective by our extensive experiments. 
Extensive experiments have been conducted to demonstrate that our overall approach is effective.
(3) In addition to video captioning, NSVA is used to study salient player identification and hierarchical action recognition. We believe this is a meaningful extension to the fine-grained action understanding domain and can help researchers gain more knowledge by investigating their sports analysis models for these new aspects. 

\section{Related work}\label{sec:related_work}
%\noindent\textbf{Natural Language Generation (NLG)} is considered as one of the fundamental problem in NLP, which aims at training models to produce coherent, understandable contents in human language for interested readers~\cite{Gatt-jair2018}. Historically, what intrigues researchers the most is the data-to-text problem that takes linguistic or non-linguistic data as input and then generates the corresponding text as output. Owing to the recent advance in the field of Deep Learning and Computer Vision, more and more attention has been paid to visual data-to-text problem that requires the model to verbally describe the content presented in an image or a short video~\cite{li-TETCI2019}.
% Jack
%\noindent\textbf{Video Captioning} aims to generate single or multiple natural language sentences based on the information stored in images or video clips. Researchers usually tackle this visual data-to-text problem with encoder-decoder framework wherein the encoder is responsible for understanding the visual signal and the decoder takes charge of generating natural language conditioned on the output of encoder. The architecture design for the encoder and decoder should take the nuance among task formulations into consideration. Traditionally in the video captioning task, 2D or 3D CNN is a popular choice for the encoder to learn the spatiotemporal feature from videos and RNN as well as its variants, e.g., LSTM and GRU, is often used in the decoder to output single or multiple coherent natural language sentence(s).
%He
\noindent\textbf{Video captioning} aims at generating single or multiple natural language sentences based on the information stored in video clips. Researchers usually tackle this visual data-to-text problem with encoder-decoder frameworks~\cite{memoryrnn,pan2020spatio,aafaq2019spatio,shi2020learning}. 
Recent efforts have found object-level visual cues particularly useful for caption generation on regular videos~\cite{pan2020spatio,objectcaption,objectcaption2,objectcaption3} as well as sports videos~\cite{SVN,svcdv}. Our work follows this idea to make use of detected finer visual features together with global information for professional sports video captioning.
%The architecture design for the encoder and decoder should take the nuance among task formulations into consideration. Traditionally in the video captioning task, 2D or 3D CNN is a popular choice for the encoder to learn the spatiotemporal feature from videos and RNN as well as its variants, e.g., LSTM and GRU, is often used in the decoder to output single or multiple coherent natural language sentence(s).

\noindent\textbf{Transformers and attention} first achieved great success in the natural language domain~\cite{attention-all-you-need,bert}, and then received much attention in vision research. One of the most influential pioneering works is the vision transformer (ViT)~\cite{vit}, which views an image as a sequence of patches on which a transformer is applied. 
%The results achieved by ViT is competitive with state-of-the-art CNNs on image classification task, which shows great potential of transformer being used in computer vision task. Shortly after ViT, 
Shortly thereafter, many tasks have found improvements using transformers, e.g., object detection~\cite{detr}, semantic segmentation~\cite{segtransformer,segtransformer2} and video understanding~\cite{tqn,swin,videobert,timeSformer}. 
Our work is motivated by these advances and uses transformers as building blocks for both feature extraction and video caption generation.
%TimeSformer~\cite{timeSformer} is one competitive model that solely builds upon self-attention blocks and achieves state-of-the-art accuracy on video action recognition. 
%ur work is motivated by these advances and try using transformer as unit building block for video captioning. To the best of our knowledge, this is the first attempt to replace CNN model with this novel transformer based architecture as feature extractor in the sport video captioning task.

\noindent\textbf{Sports video captioning} is one of several video captioning tasks that emphasizes generation of fine-grained text descriptions for sport events, e.g., chess, football, basketball and volleyball games~\cite{chen-acl2011,msrvtt,chess,svcdv,Yu-cvpr2018,SVN}. One of the biggest limitations in this area is the lack of public benchmarks. Unfortunately, none of the released video captioning datasets have a focus on sport domains.
%~\footnote{\url{https://github.com/jssprz/video\_captioning\_datasets}}. 
The most similar efforts to ours have not made their datasets publicly available~\cite{svcdv,Yu-cvpr2018,SVN}, which inspires us to take advantage of webly available data to produce a new benchmark and thereby enable  more exploration on this valuable topic. 

\noindent\textbf{Identity aware video captioning} is one of the video captioning tasks that requires recognizing person identities~\cite{nba2,nba1,identity-aware-captioning}. 
We adopt this setting in NSVA because successfully identifying players in a livestream game is crucial for sports video understanding and potential application to automatic score keeping. Unfortunately, the extant sports video captioning work failed to take player identities into consideration when creating their datasets. Earlier efforts that targeted player identification in professional sport scenes only experimented in highly controlled (i.e., unrealistic) environments, e.g., two teams and ten players, and has not consider incorporating identities in captioning~\cite{nba2,nba1}.

%jack
%\noindent\textbf{Action Recognition} task is to automate the procedure of recognizing and identifying actions in video. This task is usually formulated as a multi-class classification problem and each video only contains one action. On the contrary, video in NSVA contains variable number of actions in time order. We adopt the metrics from instructional planning literature for the evaluation of this task in NSVA. See details in Sec~\ref{sec:finegrain_action}.
% heyoma
\noindent\textbf{Action recognition} automates identification of actions in videos. Recent work has mostly focused on two sub-divisions: coarse and fine-grained recognition. The coarse level tackles basic action taxonomy and many challenging datasets are available, e.g., UCF101~\cite{ucf101}, Kinetics~\cite{kinetics} and ActivityNet~\cite{activitynet}. In contrast, fine-grained distinguishes sub-classes of basic actions, with representative datasets including Diving48~\cite{diving48}, FineGym~\cite{finegym}, Breakfast~\cite{breakfast} and  Epic-Kitchens~\cite{epickitchen}. Feature representation has advanced rapidly within the deep-learning paradigm (for review, see~\cite{acreview}) from primarily convolutional (e.g.,~\cite{tsn,i3d,s3d,tsm,slowfast}) to attention-based (e.g.,~\cite{timeSformer,swin}). Our study contributes to action understanding by providing a large-scale fine-grained basketball dataset that has three semantic levels as well as a novel attention-based recognition approach.

% \vspace{-8pt}

\section{Data collection}\label{sec:data}
Unlike previous work, we make fuller use of data that is available on the internet. We have written a webscaper to scrape NBA play-by-play data from the official website~\cite{nba}, which contains high resolution (e.g., 720P) video clips along with descriptions, each of which is a single event occurred in a game.
%\footnote{A good example for NBA play-by-play textual data is shown in this website: \url{https://www.nba.com/game/mil-vs-lal-0021900939/play-by-play}.}. 
We choose 132 games played by 10 teams in NBA season 2018-2019, the last season unaffected by COVID and when teams still could play with full capacity audiences, for data collection. We have collected 44,649 video clips, each of which has its associated play-by-play information, e.g., description, action and player names. We find that on the NBA website some different play-by-play information share the same video clip because there are multiple events taking place one-by-one within a short period time and the NBA just simply uses the same video clip for every event occurring in it. To avoid conflicting information in model training, the play-by-play text information sharing the same video clip is combined. We also remove the play-by-play text information that is beyond the scope of a single video clip, e.g., the points a player has scored so far in this game. This entire process is fully automated, so that we can access NBA webly data and associate video clips with captions, actions and players. Overall, our dataset consists of 32,019 video clips for fine-grained video captioning, action recognition and player identification. Additional details on dataset curation are provided in the supplement.

\subsection{Dataset statistics}
\begin{table*}[t]
% \scriptsize
\centering
\resizebox{\columnwidth}{!}{%
\begin{tabular}{l c c c cc c c c c c c c c c c c}
\toprule
Datasets & & Domain && \texttt{\#}Videos & & \texttt{\#}Sentences & & \texttt{\#}Hours & & Avg. words & & Accessibility & & Scalability & & Multi-task \\
\midrule
SVN~\cite{SVN} & & basketball & & 5,903 & & 9,623 & & 7.7 & & 8.8 & & \ding{55} & & \ding{55}& & \ding{55}  \\
SVCDV~\cite{svcdv} & & volleyball & & 4,803 & & 44,436 & & 36.7 & & - & &\ding{55} & & \ding{55} & & \ding{55} \\
NSVA & & basketball & & \textbf{32,019} && \textbf{44,649} & &\textbf{84.8} & & 6.5 & & \ding{51} & & \ding{51} & & \ding{51} \\ 
\bottomrule
\end{tabular}
}
\caption{The statistics of NSVA and comparison to other fine-grained sports video captioning datasets.
}
%\textcolor{blue}{include the basket-subdomain from ActivityNet}}
\label{tab:stats}
% \vspace{-15pt}
\end{table*}
\begin{table*} [b]
\resizebox{\columnwidth}{!}{%
\centering
    % \resizebox{0.8\textwidth}{!}{
        \begin{tabular}{*{4}{c} c *{4}{c} c *{4}{c} c c c c c c}
        \toprule
        \multicolumn{4}{c}{Videos} &
                               & \multicolumn{4}{c}{Sentences} &
                               & \multicolumn{4}{c}{Games} & &
                               Teams & &
                               Actions & &
                               Identities
                               \\
                               \cmidrule(l){1-4} 
                                \cmidrule(l){5-9} 
                                \cmidrule(l){10-14} 
                                \cmidrule(l){15-16}
                                \cmidrule(l){17-18}
                                \cmidrule(l){19-20}
                               train & val & test & total &
                               & train & val & test & total  &
                               & train & val & test & total & &
                               all-sets & &
                               all-sets & &
                               all-sets \\
            24k & 3.9k & 3.9k & 32k & & 33.6k & 5.5k & 5.5k & 44.6k & & 100 & 16 & 16 & 132 & & 10 & & 172 & & 184 \\
        \bottomrule
    \end{tabular}
    }
    % }
    \caption{Data split detail of our dataset.}
    \label{tab:data_split}
    % \vspace{-21pt}
\end{table*}
Table~\ref{tab:stats} shows the statistics of NSVA and two other fine-grained sports video captioning datasets. NSVA has the most sentences out of three datasets and five times more videos than both SVN and SVCDV. The biggest strength of NSVA is its public accessibility and scalability. Both SVN and SVCDV datasets are neither publicly available nor scalable because heavy maunal annotation effort is required in their creation. In contrast,  NSVA is built on data that already existed on the internet; so, everyone who is interested can directly download and use the data by following our guidelines. Indeed, the 132 games that we chose to use only accounts for 10.7\% of total games in NBA season 2018-2019. There is more data being produced everyday as NBA teams keep playing and sharing their data. Note that some other datasets also contain basketball videos, e.g., MSR-VTT~\cite{msrvtt} and ActivityNet~\cite{activitynet}. However, they only provide coarse-level captions (see example in Figure~\ref{fig:teaser}) and include very limited numbers of videos, e.g, ActivityNet has 74 videos for basketball and they are all from amateur play, not professional. 

Table~\ref{tab:data_split} shows the data split of NSVA. We hold 32 games out from 132 games to form validation set and test set, each of which contains 16 games. All clips and texts belonging to a single game are assigned to the same data split. When choosing what data split a game is assigned to, we ensure that every team match-up has been seen at least once in the training set. For example, Phoenix Suns play four games against San Antonio Spurs in NBA season 2018-2019. We put two games in the training set, one in the validation set and one in the test set.

NSVA also supports two additional vision tasks, namely fine-grained action recognition and key player identification. We adopt the same data curation strategy as captioning and show the number of distinct action or player name categories in the rightmost two columns of Table~\ref{tab:data_split}. When being compared with other find-grained sport action recognition datasets, e.g., Diving48 (48 categories) and Finegym (530 categories), ours is in the middle place (172 categories) in terms of number of actions and is the largest regarding the basketball sub-domain. 
%We This constitutes another advantage of NSVA. Researchers can understand their model’s performance better by evaluating it on these two subtasks.
%\subsection{Line numbering}
\begin{figure}[t]
\footnotesize
\centering
\includegraphics[width=1.0\linewidth]{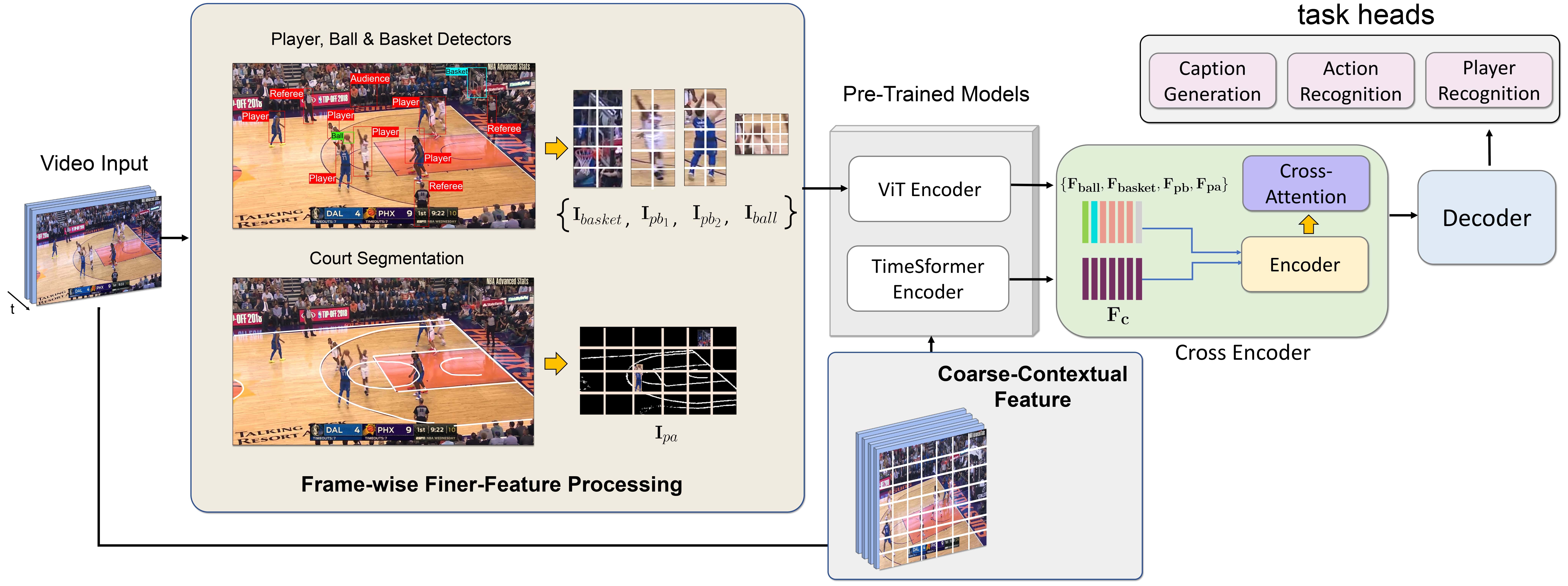}
\caption{Pipeline of our proposed approach for versatile sports video understanding. First, raw video clips (left) are processed into two types of finer visual information, namely object detection (including ball, players and basket), and court-line segmentation, all of which are cropped, grided and channelled into a pre-trained vision transformer model for feature extraction. Second, these heterogeneous features are aggregated and cross-encoded with the global contextual video representation extracted from TimeSformer (middle). Third, a transformer decoder is used with task-specific heads to recursively yield results, be it as video captions, action recognition or player identification (right). 
}
\label{fig:algorithm}
\end{figure}

% All lines should be numbered in the initial submission, as in this example document. This makes reviewing more efficient, because reviewers can refer to a line on a page. Line numbering is removed in the camera-ready.
\section{Architecture design}\label{sec:methods}
\noindent\textbf{Problem formulation.} We seek to predict the correct sequence of word captions as one-hot vectors, $\{\mathbf{y}\}$, whose length is arbitrary, given the observed input clip $X \in \mathbb{R}^{H \times W \times 3 \times N}$ consisting of $N$ RGB frames of size $H \times W$ sampled from the original video.

\noindent\textbf{Overall structure}. As our approach relies on feature representations extracted from multiple orthogonal perspectives, we adopt the framework of UniVL \cite{UniVL}, a network designed for cross feature interactive modeling, as our base model. It consists of four transformer backbones that are responsible for coarse feature encoding, fine-grained feature encoding, cross attention and decoding, respectively. 
In the following, we step-by-step detail our multi-level feature extraction, integrated feature modeling and decoder.

\subsection{Course contextual video modeling}
%The only convolutional part of UniVL is its video feature extractor which uses S3D model to learn spatiotemporal representation of video. Inspired by TimeSformer which is solely built on transformer block and has shown great performance on several action recognition datasets, we substitute the S3D part of UniVL with this new model as video feature extractor thus making our model convolution-free. Compared to S3D which uses CNN kernels that lay stress on the locality and short-range spatiotemporal information, TimeSformer builds on Transformer block that has less strong inductive priors so it can easily model long-range spatiotemporal information with its self-attention mechanism in a large-scale data learning setting. To the best of our knowledge, this is the first attempt that brings the TimeSformer from action recognition regime to video captioning.
In most video captioning efforts a 3D-CNN has been adopted as the fundamental unit for feature extraction, e.g., S3D \cite{s3d,SVN,svcdv}. More recent work employed a transformer architecture in tandem \cite{UniVL}.
%The only convolutional part of UniVL is its video feature extractor which uses S3D model to learn spatiotemporal representation of video. 
Inspired by TimeSformer \cite{timeSformer}, which is solely built on a transformer block and has shown strong performance on several action recognition datasets, we substitute the S3D part of UniVL with this new model as video feature extractor. Correspondingly, we decompose each frame into $F$ non-overlapping patches, each of size $P \times P$, such that the $F$ patches span the entire frame, i.e., $F = HW/P^2$.
We flatten these patches into vectors and channel them into several blocks comprised of  linear-projection, multihead-self-attention and layer-normalization, in both spatial and temporal axes, which we shorten as
\begin{equation}\label{eq:coarse}
    \mathbf{F}_c = \text{TimeSformer} \left(X\right),
\end{equation}
where $\mathbf{F}_c \in \mathbb{R}^{N \times d}$, $d$ is the feature dimension and $X$ is an input clip.

%Compared to S3D, which uses CNN kernels that lay stress on the locality and short-range spatiotemporal information, TimeSformer builds on a transformer block that has less strong inductive priors so it can easily model long-range spatiotemporal information with its self-attention mechanism in a large-scale data learning setting. To the best of our knowledge, this is the first attempt that brings the TimeSformer from the action recognition regime to video captioning.

Transformer blocks have less strong inductive priors compared to convolutional blocks, so they can more readily model long-range spatiotemporal information with their self-attention mechanism in a large-scale data learning setting. We demonstrate the strong performance of TimeSformer features in Sec.~\ref{sec:experiment}.

\subsection{Fine-grained objects of interest modeling}
One limitation of solely using TimeSformer features is that we might lose important visual details, e.g., ball, players and basket, after resizing $1280\times720$ images to $224\times224$, the size that TimeSformer encoder needs. Such loss can be important because NSVA requires modeling main players' identities and their actions to generate an accurate caption. To remedy this issue, we use an object detector to capture objects of interest that contain rich regional semantic information complementary to the global semantic feature provided by TimeSformer. We extract 1,000 image frames from videos in the training set and annotate bounding boxes for basket and ball and fine-tune on the YOLOv5 model \cite{yolov5} to have a joint ball-basket object detector. This pre-trained model returns ball and basket crops from original images, i.e., $\mathbf{I}_{ball}$ and $\mathbf{I}_{basket}$.  

For player detector, we simply use the YOLOv5 model trained on the MS-COCO dataset~\cite{ms-coco} to retrieve a stack of player crops, $\{ \mathbf{I}_{player}\}$. As our caption is identity-aware, we assume that players who have touched the ball during a single play are more likely to be mentioned in captions. Thus, we only keep the detected players that have overlap with a detected ball, e.g., each player crop, $\mathbf{I}_{player}$, is given a confidence score, $C$, of 1 otherwise 0; in particular, $\text{if} \; \text{IoU}\left(\mathbf{I}_{player_{i}}, \mathbf{I}_{ball}\right) > 0:C = 1; \; \text{else}: C = 0$. Player crops that have $C = 1$ will be selected for later use, $\mathbf{I}_{pb}$. Even though the initially detected players, $\{\mathbf{I}_{player}\}$, potentially are contaminated by non--players (e.g., referees, audience members), our ball-focused confidence scores tend to filter out these distractors.

After getting bounding boxes of ball, players intersecting with the ball and basket, we crop these objects from images and feed them to a vision transformer, ViT~\cite{vit}, for feature extraction,
% \begin{equation}
% \mathbf{F_{ball}}=\text { ViT }\left(\mathbf{i}_{ball}\right), \mathbf{F_{pb}}=\text {ViT}\left(\mathbf{i}_{pb}\right), 
% \mathbf{F_{basket}}=\text {ViT}\left(\mathbf{i}_{basket}\right)
% \end{equation}
\begin{equation}
    \mathbf{f}_{ball} = \text{ViT}\left(\mathbf{I}_{ball}\right), \;  \mathbf{f}_{basket} = \text{ViT}\left(\mathbf{I}_{basket}\right), \; 
    \mathbf{f}_{pb} = \text{ViT}\left(\mathbf{I}_{pb}\right),
\end{equation}
where $\mathbf{f}_{ball}$, $\mathbf{f}_{pb}$ and $\mathbf{f}_{basket}$ are features of $d$ dimension extracted from cropped ball image, $\mathbf{I}_{ball}$, player with ball image, $\mathbf{I}_{pb}$, and basket image, $\mathbf{I}_{basket}$, respectively. We re-group features from every second in the correct time order to have $\mathbf{F}_{ball}$, $\mathbf{F}_{basket}$ and $\mathbf{F}_{pb}$, which all are of dimensions $\mathbb{R}^{m \times d}$.

\noindent\textbf{Discussion.} Compared with previous work that either require pixel-level annotation in each frame to segment each player, ball and background~\cite{SVN}, or person-level annotation that needs professional sport knowledge to recognize each player's action such as setting, spiking and blocking~\cite{svcdv}, our annotation scheme is very lightweight. The annotation only took two annotators less than five hours to draw bounding boxes for ball and basket in 1,000 selected image frames from the training set. Compared to the annotation procedure that requires months of work for experts with extensive basketball knowledge~\cite{SVN}, our approach provides a more affordable, replicable and scalable option. Note that these annotations are only for training the detectors; the generation of the dataset per se is completely automated; see Sec.~\ref{sec:data}.
%for other researchers interested in this task.

\subsection{Position-aware module}

NSVA supports modeling estimation of the distance from where the main player's actions take place to the basket. As examples, ``Lonnie Walker missed \textbf{2'} cutting layup shot'' and ``Canaan \textbf{26'} 3PT Pullup Jump Shot'', where the numbers in bold denote the distance between the player and basket. 
Notably, distance is strongly correlated with action; e.g., players cannot make a 3PT shot at two-foot distance from the basket.
While estimating such distances is important for action recognition and caption generation, it is non-trivial owing to the need to estimate separation between two 3D objects from their 2D image projections. 

Instead of explicitly making such prediction directly on raw video frames, we take advantage of prior knowledge that basketball courtlines are indicators of object's location. We use a pix2pix network \cite{pix2pix} trained on synthetic data \cite{reconstructnba} to generate courtline segmentation given images.
%\footnote{\url{https://github.com/luyangzhu/NBA-Players}}. 
We overlay the detected player with ball and basket region, while blacking out other areas. Figure~\ref{fig:algorithm} shows an exemplar image, $\mathbf{I}_{pa}$, after such processing. We feed these processed images to ViT for feature extraction, i.e.,
$\mathbf{F_{pa}}=\text { ViT }\left(\mathbf{I}_{pa}\right),
$
where $\mathbf{F}_{pa} \in \mathbb{R}^{m \times d}$ are ViT features extracted from position-aware image $\mathbf{I}_{pa}$.

% Note that in our work we do not use optical flow, cf.,~\cite{SVN}, which has proven effective in other similar work~\cite{SVN,svcdv} because we do not have annotations to distinguish the moving objects, e.g., ball, judges, players and still objects, e.g., audience and basket, 
% %\footnote{It is possible to estimate camera speed using detected basket but basket does not always exist in our case, e,g, no basket can be seen when a player is running through the half court line}, 
% which is needed for estimating camera speed and producing an unnoisy optical flow.  

\subsection{Visual transformer encoder}
After harvesting the video, ball, basket and courtline features, we are ready to feed them into the coarse encoder as well as the finer encoder for self-attention. This step is necessary as the used backbones (i.e., ViT and TimeSformer) only perform attention on frames within one second; there is no communication between different timestamps.
%This process is to encode the contextual information of video features.
For this purpose, we use one transformer to encode video feature, $\mathbf{F}_{c} \in \mathbb{R}^{N \times d}$ \eqref{eq:coarse}, and another transformer to encode aggregated finer features, $\mathbf{F}_f \in \mathbb{R}^{M \times 2d}$, which is from the concatenation of position-aware feature, $\mathbf{F}_{pa}$, and the summation of object-level features. Empirically, we find summation sufficient, i.e., 
\begin{equation}
    \mathbf{F}_{f} = \text {CONCAT}(\text {SUM}( \mathbf{F}_{ball}, \mathbf{F}_{basket},\mathbf{F}_{pb}),\mathbf{F}_{pa})
\end{equation}

The overall encoding process is given as
\begin{equation}
    \mathbf{V}_{c} = \text{Transformer}\left(\mathbf{F}_{c}\right), 
    \mathbf{V}_{f} = \text{Transformer}\left(\mathbf{F}_{f}\right),
\end{equation}
where $\mathbf{V}_{c} \in \mathbb{R}^{n \times d}$ and $\mathbf{V}_{f} \in \mathbb{R}^{m \times d}$.
%\footnote{Usually we have $m < n$ for saving storage space. See Sec.~\ref{subsec:Implementation details} for more details.} 

\subsection{Cross encoder for feature fusion}
The coarse and fine encoders mainly focus on separate information. To make them fully interact, we follow existing work and adopt a cross encoder \cite{UniVL}, which takes coarse features, $\mathbf{V}_{c}$, and fine features, $\mathbf{V}_{f}$, as input. Specifically, these features are combined along the sequence dimension via concatenation and a transformer is used to generate the joint representation, i.e.,
%Cross encoder is used to fuse global contextual feature ${V_c}$ and multi-aspect feature ${V_f}$,
\begin{equation}
\mathbf{M}=\operatorname{Transformer}(\text {CONCAT}(\mathbf{V_c}, \mathbf{V_f})),
\end{equation}
where $\mathbf{M}$ is the final output of the encoder. To generate a caption, a transformer decoder is used to attend $\mathbf{M}$ and output text autoregressively, cf., \cite{videobert,gpt3,radford2021learning}. 

\subsection{Learning and inference}
Finally, we calculate the loss as the sum of negative log likelihood of correct caption at each step according to
% \begin{equation}
% L_{c a p}=-\sum_{t=1}^{T} \log p\left(y_{t} \mid v_{g}, x_{t}, \mathcal{B}_{t}, \theta^{*}\right)
% \end{equation}
\begin{equation}\label{eq:loss}
\mathcal{L}(\theta)=-\sum_{t=1}^{T} \log P_{\theta}\left({y}_{t} \mid {y}_{<t}, \mathbf{M}\right),
\end{equation}
where $\theta$ is the trainable parameters, ${y}_{<t}$ is the ground-truth words sequence before step $t$ and $y_{t}$ is the ground truth word at step $t$.

During inference, the decoder autoregressively operates a beam search algorithm~\cite{beam} to produce results, with beam size set empirically; see Sec.~\ref{subsec:Implementation details}.

\subsection{Adaption to other tasks}
In NSVA, action and identity also are sequential data. So, we adopt the same model, shown in Figure~\ref{fig:algorithm}, for all three tasks and swap the caption supervision signal in \eqref{eq:loss}, $y_{1:t}$, with either one-hot action labels or player name labels. Similarly, inference operates beam search decoding. Details are in the supplement.

\section{Empirical evaluation}\label{sec:experiment}
\subsection{Implementation details}\label{subsec:Implementation details}
We use hidden state dimension of 768 for all encoders/decoders. We use the BERT~\cite{bert} vocabulary augmented with 356 action types and player names entries. The transformer encoder, cross-attention and decoder are pre\-trained on a large instructional video dataset, Howto100M~\cite{howto100m}. We keep the pre-trained model and fine tune it on NSVA, as we found the pre-trained weights speed up model convergence. The maximum number of frames for the encoder and the maximum output length are set to 30. The number of layers in the feature encoder, cross encoder and decoder are 6, 3 and 3, respectively. We use the Adam optimizer with an initial learning rate of  3e-5 and employ a linear decay learning rate schedule with a warm-up strategy. We used a batch size of 32 and trained our model on a single Nvidia Tesla T4 GPU for 12 epochs over 6 hours. The hyperparameters were chosen based on the top performer on the validation set. 

In testing we adopt beam search~\cite{beam} with beam size 5. For extraction of the TimeSformer feature, we sample video frames at 8 fps. For extraction of other features, we sample at 12 vs 4 fps when the ball is vs is not detected in the basket area. We record the time when the ball first is detected and keep 100 frames before and after. This step saves about 70\% storage space compared to sampling 
the entire video at 8 fps, but still keeps the most important frames. 
\begin{table}[t]
\centering
\resizebox{\columnwidth}{!}{%
\begin{tabular}{l c l c c c c c c c}
\toprule
Model & & Feature & C & M & B@1 & B@2 & B@3 & B@4 & R\_L      \\
\midrule
MP-LSTM~\cite{MP-LSTM}  &   & S3D            & 0.500          & 0.153          & 0.325          & 0.236          & 0.167          & 0.121          & 0.332          \\ 
TA~\cite{TA}        &  & S3D            & 0.546          & 0.156          & 0.331          & 0.242          & 0.175          & 0.128          & 0.340          \\
Transformer~\cite{Transformer} & & S3D            & 0.572          & 0.161          & 0.346          & 0.254          & 0.181          & 0.131          & 0.357          \\
UniVL$^{*}$~\cite{UniVL}   &    & S3D            & 0.717          & 0.192          & 0.441          & 0.309          & 0.226          & 0.169          & 0.401          \\ 
\midrule
       & & T              & 0.956          & 0.217          & 0.467          & 0.363          & 0.274          & 0.209          & 0.468          \\
       & & S3D+BAL+BAS+PB+PA & 0.986          & 0.227          & 0.479          & 0.371          & 0.281          & 0.216          & 0.466          \\
       & & T+BAL          & 0.931          & 0.228          & 0.496          & 0.383          & 0.289          & 0.220         & 0.484          \\
       & & T+BAS          & 1.023          & 0.232          & 0.500          & 0.387          & 0.292          & 0.223          & 0.486          \\
\multicolumn{1}{l}{Our Model}       & & T+PB           & 1.055          & 0.231          & 0.500          & 0.387          & 0.292          & 0.223          & 0.487          \\
       & & T+PA           & 1.064 & 0.238 & 0.511 & 0.398 & 0.301 & 0.231 & 0.498 \\
       & & T+BAL+BAS      & 1.074          & 0.243          & 0.508          & 0.398          & 0.306          & 0.237          & 0.499          \\
       & & T+BAL+BAS+PB   & 1.096          & 0.242          & 0.519          & 0.408          & 0.312          & 0.242          & 0.506          \\
       & & T+BAS+BAL+PB+PA    & \textbf{1.139} & \textbf{0.243} & \textbf{0.522} & \textbf{0.410} & \textbf{0.314} & \textbf{0.243} & \textbf{0.508}
       \\
      \bottomrule
\end{tabular}}
\caption{Performance comparison of our model vs. alternative video captioning models on the NSVA test set. T denotes TimeSformer feature. BAL, BAS and PB denote ViT features for ball, basket and player with ball, respectively. PA is the position-aware feature. $^{*}$As our model adopts the framework of UniVL as backbone, results in the row of UniVL+S3D equals to those of our model only using S3D features.}
\label{tab:caption_rst}
\end{table}

\subsection{Video captioning}
\noindent\textbf{Baseline and evaluation metrics.} The main task of NSVA is video captioning. To assess our proposed approach, we compare our results with four state-of-the-art video captioning systems: MP-LSTM~\cite{MP-LSTM}, TA~\cite{TA}, Transformer~\cite{Transformer} and UniVL~\cite{UniVL} on four widely-used evaluation metrics: CIDEr (C) \cite{cider}, Bleu (B) \cite{bleu}, Meteor (M) \cite{meteor} and Rouge-L (R\_L) \cite{rouge}. Results are shown in Table~\ref{tab:caption_rst}. 
%\textcolor{blue}{how do we re-implement comparison baselines on our own dataset? need to write x-modelar link.} 
To demonstrate the effectiveness of our approach against the alternatives, we train these models on NSVA using existing codebases~\cite{Xmodaler,UniVL}.
%\footnote{X-modaler: A Versatile and High-performance Codebase
%for Cross-modal Analytics, \url{https://github.com/YehLi/xmodaler}.}

\noindent\textbf{Main results.} Comparing results in the first two rows with results in other rows of Table~\ref{tab:caption_rst}, we see that transformer models outperform LSTM models, which confirms the superior capability of a transformer on the video captioning task. Moveover, it is seen that TimeSformer features achieve much better results compared to S3D in modeling video context. We conjecture that this is due to its ability to model long spatiotemporal dependency in videos; see $4^{th}$ and $5^{th}$ rows. This result suggests that TimeSformer features are not only useful for video understanding tasks but also video captioning. Comparing results on the $4^{th}$ and $6^{th}$ rows, we find that after fusing S3D features with those extracted by our proposed modules (but not the TimeSformer), improvements are seen on all metrics.
%This shows that our proposed finer-feature module can be very compatible with varied features. 
A possible explanation is that our features add additional semantic information (e.g., pertaining to ball, player and court) and thereby lead to higher quality text. The best result is achieved by fusing TimeSformer features with our proposed features. These results suggest that (1) TimeSformer features are well suited to video captioning and (2) our proposed features can be fused with a variety of features for video understanding to improve performance further. 
\begin{comment}
\noindent\textbf{Meteor results.} It is seen that PA and PB features are individually effective on the Meteor metric, e.g., leading to notable improvements over TimeSformer feature alone. However, a plateau occurs when using PA and PB with others, i.e., T+BAL+BAS+PB+PA. Possible explanations are: (1) The combination of features has saturated the dataset. (2) In its formal definition~\cite{meteor}, the Meteor score is defined as $ F_{mean} \times (1-\text{Penalty})$. The second term, i.e., $(1-\text{Penalty}) \in [0,5, 1)$, therefore discounts the performance improvements. (3) The Meteor score increases most when evaluating captions that have identical stem (e.g., missing vs. missed) as well as synonym (e.g., passing vs hand-over) with the ground truth. The presence of such ambiguities in NSVA is very minor. 
\end{comment}
\begin{figure}[t]
\centering
\includegraphics[width=1.0\linewidth]{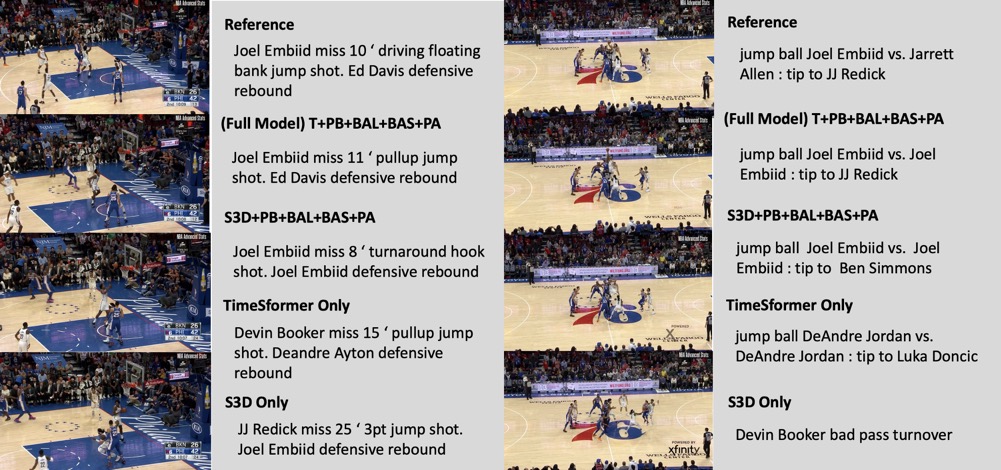}
\caption{Qualitative analysis of captions generated by our proposed approach and others. It is seen that captions from our full approach are the most close to references.
}
\label{fig:qualitative_analysis}
% \vspace{-12pt}
\end{figure}
From the $7^{th}$ to final row of Table~\ref{tab:caption_rst}, we ablate our finer-grained features. It is seen that our model benefits from every proposed finer module, and when combining all modules, we observe the best result; see last row. This documents the effectiveness of our proposed method for the video captioning task on NSVA. More discussions on the empirical results can be found in the supplement.

\noindent\textbf{Qualitative analysis.} Figure~\ref{fig:qualitative_analysis} shows two example outputs generated by four different models, as compared to the ground-truth reference. From the left example output, we see that our full model is able to generate a high quality caption, albeit with relatively minor mistakes. After replacing the TimeSformer features with S3D features, the model fails to identify the player who gets the rebound and mistakes a jump shot for a hook shot. When using TimeSformer or S3D feature alone, the result further deteriorate by misidentifying all players. We also notice that our devised features, i.e., PB+BAL+BAS+PA, can greatly help capture a player's position, e.g., with 10' as the reference, models with PB+BAL+BAS+PA features output 11' and 8', compared to 15' and 25' output by TimeSformer only and S3D only. 

The right column shows an example where all models successfully recognize the action, i.e., jump shot, except the S3D only model. Our full model can identify most players but still mistakes Jarret Allen for Joel Embiid. As we will discuss in Sec.~\ref{sec:playerID}, player identification is the bottleneck of our model as it is trained with a very weak supervision signal, which points to future research. %We leave the solution to this problem as future work.

\subsection{Fine-grained basketball action recognition} \label{sec:finegrain_action}
\begin{figure}[t]
\centering
\includegraphics[width=1.0\linewidth]{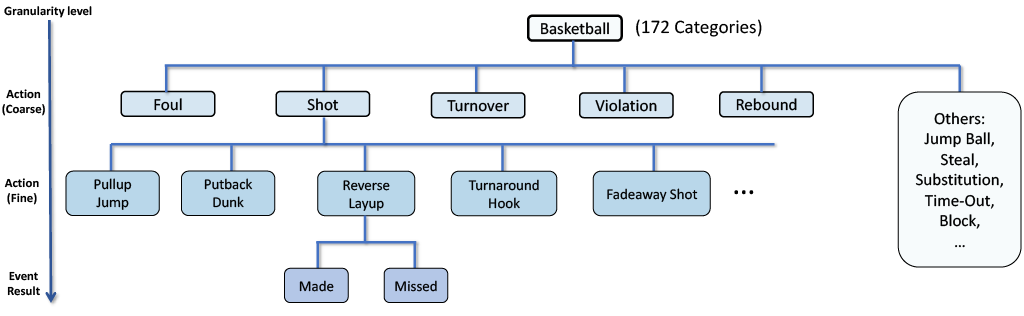}
\caption{
Visualization of a sub-tree from our fine-grained basketball action space. 
There are 172 fine-grained categories that comprise three levels of sport event details: \textbf{Action-C} (coarse), \textbf{Action-F} (fine) and \textbf{Action-E} (event). Some categories have finer descendants (e.g., \textit{Shot}), while others are solitary (e.g., \textit{Jump Ball} and \textit{Block}). The full list of action categories is in the supplement.
}
\label{fig:action_categories}
\end{figure}

As elaborated in Sec.~\ref{sec:data}, NSVA has massive video clips that cover almost every moment of interest, and these events have been provided by the NBA for the purpose of statistics tracking, which allows fine-grained action recognition. A glimpse of how our action labels are hierarchically organized is shown at Figure~\ref{fig:action_categories}.

\noindent\textbf{Action hierarchy.} NSVA enjoys three levels of granularity in the basketball action domain. (1) On the coarsest level, there exist 14 actions that describe the on-going sport events from a very basic perspective. Some representative examples include: $\{$ \textcolor{green}{\textit{Shot, Foul, Turnover}} $\}$. (2) If further dividing the coarse actions into their finer sub-divisions, we can curate 124 fine-grained actions. Taking the shot category as an example, it has the following sub-categories: $\{$ \textcolor{cyan}{\textit{Shot Dunk}}, \textcolor{cyan}{\emph{Shot Pullup Jumpshot}}, \textcolor{cyan}{\textit{Shot Reverse Layup}}, etc. $\}$. All of these finer actions enrich the coarse ones with informative details (e.g., diverse styles for the same basketball movement). (3) On the finest level, there exists 24 categories that depicts the overall action from the event perspective, which includes the coarse action name, the fine action style and the overall event result, e.g., $\{$ \textcolor{red}{\textit{Shot-Pullup-Jumpshot-Missed}} $\}$. Thanks to the structured labelling, NSVA can support video action understanding on multiple granularity levels. We demonstrate some preliminary results using our proposed approach in Table~\ref{tab:action_reco}.

\noindent\textbf{Evaluation.} As exemplified in Figure~\ref{fig:teaser}, our action labels do not always assign a single ground-truth label to a clip. In fact, they contain as many actions as happens within the length of a unit clip. The example in Figure~\ref{fig:teaser} shows a video clip that has two consecutive actions, i.e., \textbf{[} \textcolor{red}{\textit{3-pt Jump-Shot Missed}} $\rightarrow$ \textcolor{cyan}{\textit{Defensive Rebound}} \textbf{]}. To properly evaluate our results in this light, we adopt metrics from efforts studying instructional videos \cite{procedureplan1,pp2,pp3}, and report: (1) mean Intersection over Union (mIoU), (2) mean Accuracy (Acc.) and (3) Success Rate (SR). Detailed explanation can be found in the supplement. We provide action recognition results using the same feature design introduced in Sec.~\ref{sec:methods} and provide an ablation study on the used features.
%Following previous , we evaluate the performance using three increasingly strict metrics. (i) mean Intersection over Union (mIoU) treats the predicted and ground truth action sequences as sets, and measures the overlap between these sets. mIoU is agnostic to the order of actions and only indicates whether the model captures the correct set of steps needed to complete the plan. (ii) mean Accuracy (Acc.) performs element-wise comparisons between the predicted and ground truth action sequences, thereby considering the order of the actions as well. (iii) Success Rate (SR) considers a recognition successful only if it exactly matches the ground truth.
%work~\cite{procedureplan1, pp2}
\begin{table}[t]
\centering
\resizebox{\columnwidth}{!}{%
\begin{tabular}{c c c c c c c c c c c c c c c c}
\toprule
& & & & & & & \multicolumn{3}{c}{Action-C} & \multicolumn{3}{c}{Action-F} & \multicolumn{3}{c}{Action-E} \\
\cmidrule(l){8-10} 
\cmidrule(l){11-13} 
\cmidrule(l){14-16} 
Feature-backbone & & PB & BAL & BAS & PA & &  SR$\uparrow$ & Acc.$\uparrow$ & mIoU$\uparrow$ & SR$\uparrow$ & Acc.$\uparrow$ & mIoU$\uparrow$  &  SR$\uparrow$ & Acc.$\uparrow$ & mIoU$\uparrow$\\ 
\midrule
TimeSformer  &  & \ding{51} & \ding{51} &\ding{51} & \ding{51} &  & \textbf{60.14} & \textbf{61.20} & \textbf{66.61} & \textbf{46.88} & \textbf{51.25} & 57.08 & \textbf{37.67} & \textbf{42.34} & \textbf{46.45}\\ 
TimeSformer  &  & \ding{51} & \ding{51} &\ding{51} & - &  & 60.02 & 60.79 & 65.33 & 46.42 & 50.64 & \textbf{57.19} & 36.44 & 42.29 & 42.14\\ 
TimeSformer  &  & \ding{51} & \ding{51} & - & - &  & 58.06 & 60.31 & 63.71 & 44.31 & 49.01 & 55.78  & 34.53 & 39.34 & 46.45\\ 
TimeSformer  &  & \ding{51} & - & - & - & & 57.74 & 58.13 & 60.48 & 44.20 & 50.18 & 55.91 & 34.50 & 39.14 & 42.72 \\
TimeSformer & & - & - & - & - & & 55.83 & 58.01 & 60.19 & 42.55 & 49.66 & 53.81 & 33.63 & 37.50 & 40.84\\
S3D & & - & - & - & - & & 54.46 & 57.91 & 59.91 & 41.92 & 48.81 & 53.77 & 33.09 & 37.11 & 40.77\\
\bottomrule
\end{tabular}
}
\caption{Action recognition accuracy ($\%$) on NSVA at all granularities.}
\label{tab:action_reco}
\end{table}

\noindent\textbf{Results on multiple granularity recognition.} From the results in Table~\ref{tab:action_reco}, we can summarize several observations: (1) Overall, actions in NSVA are quite challenging to recognize, as the best result on the coarsest level only achieves $61.2\%$ accuracy (see columns under Action-C). (2) When the action space is further divided into sub-actions, the performance becomes even weaker (e.g., 51.25$\%$ for Action-F and 42.43 $\%$ for Action-E), meaning that subtle and challenging differences can lead to large drops in recognizing our actions. (3) TimeSformer features perform better than S3D counterparts at all granularity levels, which suggests NSVA benefits from long-term modeling. (4) We observe solid improvements by gradually incorporating our devised finer features, which once again demonstrates the utility of our proposed approach. 

\subsection{Player identification}\label{sec:playerID}
\begin{figure}[h]
    \centering
    \begin{minipage}[b]{1.0\textwidth}
    \centering
    \includegraphics[width=0.32\textwidth]{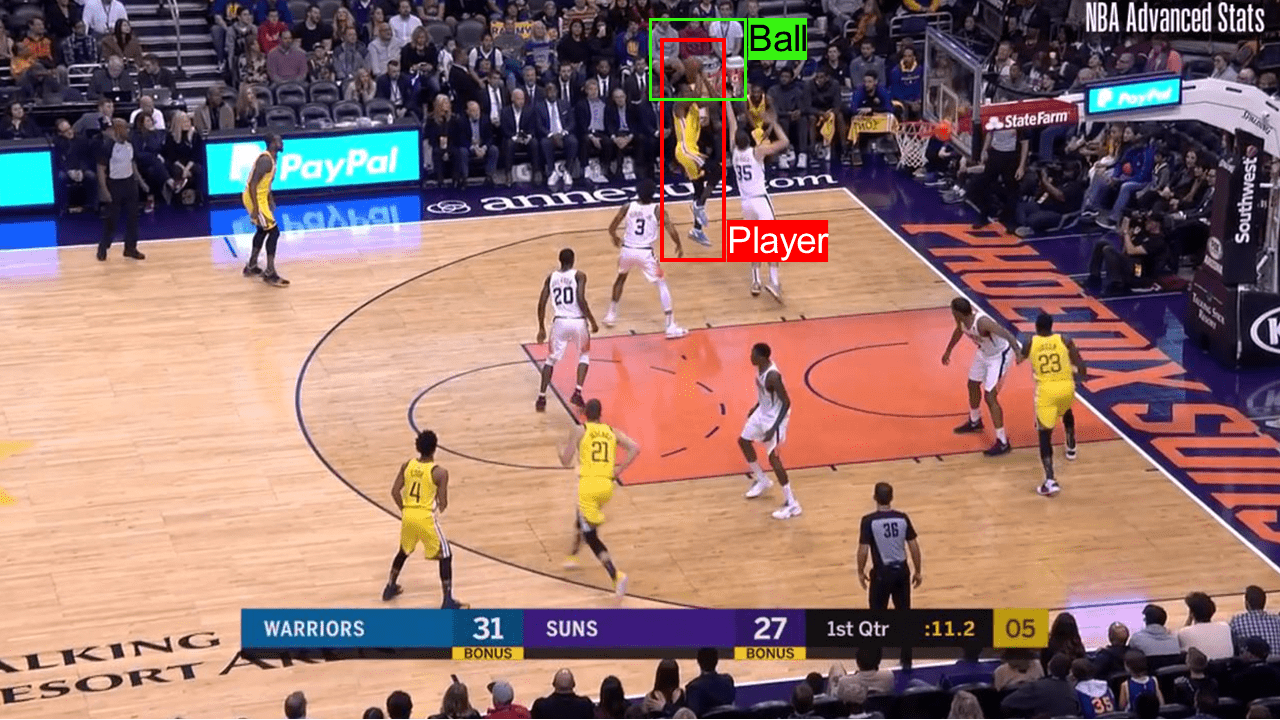}
    \includegraphics[width=0.32\textwidth]{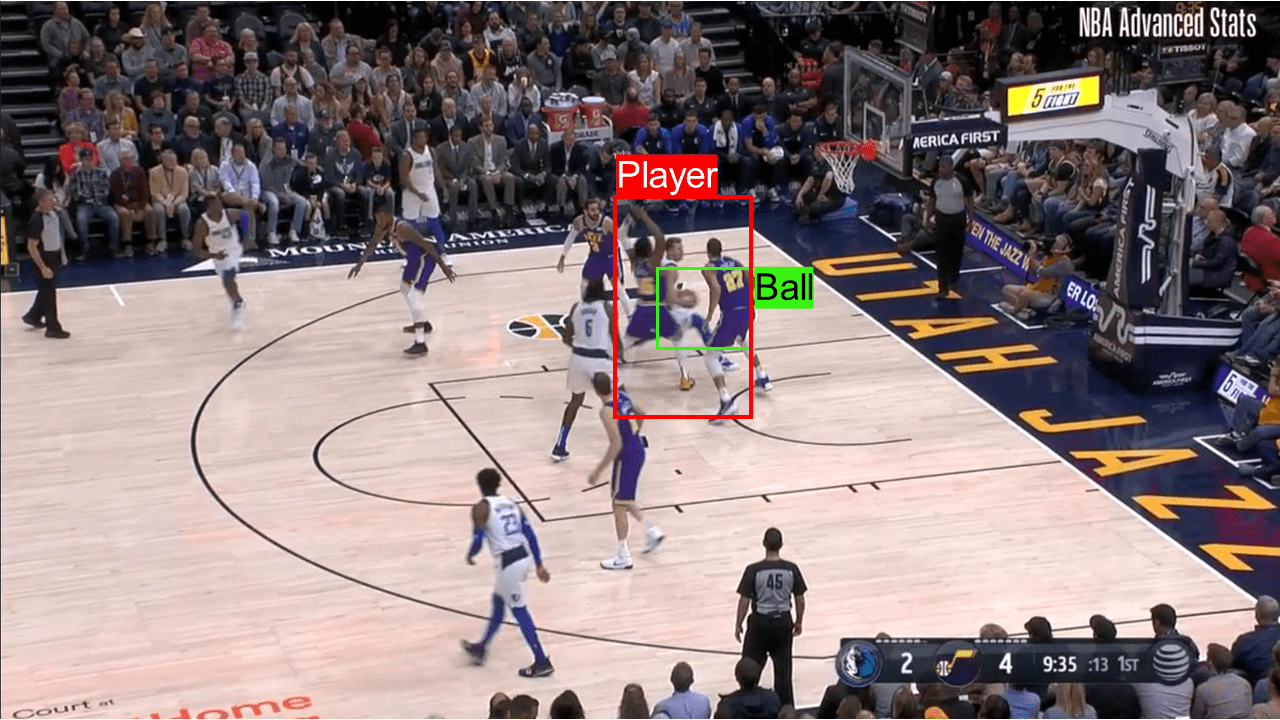}
    \includegraphics[width=0.32\textwidth]{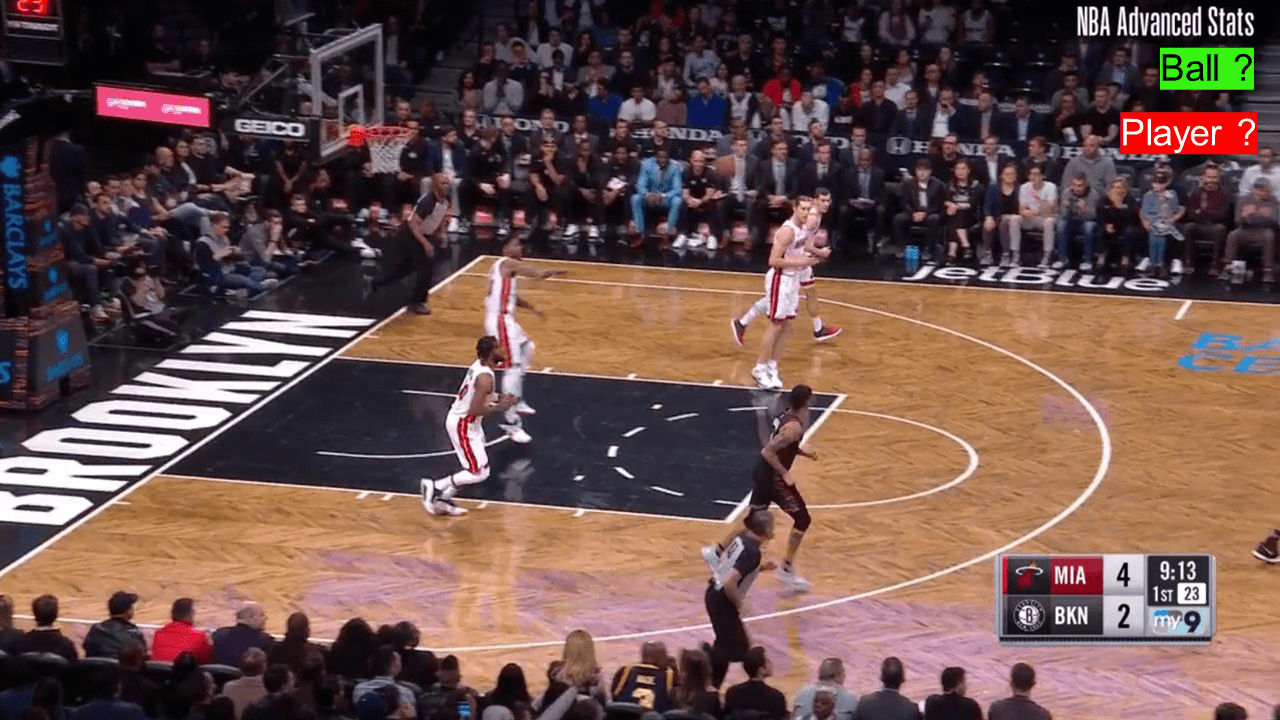}
    \label{fig:fig15}
    \end{minipage}
    \begin{minipage}[b]{.6\textwidth}
    \centering
    \scriptsize
    \begin{tabular}{c c c c c c c c c c}
        \toprule
        Feature-backbone & & PB & BAL & BAS & PA & &  SR $\uparrow$ & Acc $\uparrow$ & mIoU $\uparrow$ \\ 
        \midrule
        TimeSformer  &  & \ding{51} & \ding{51} &\ding{51} & \ding{51} &  & \textbf{4.63} & \textbf{6.97} & 6.86 \\ 
        TimeSformer  &  & \ding{51} & \ding{51} &\ding{51} & - &  & 4.20 & 6.83  & \textbf{6.89}\\ 
        TimeSformer  &  & \ding{51} & \ding{51} & - & - &  & 4.17 & 6.45 & 6.68 \\ 
        TimeSformer  &  & \ding{51} & - & - & - &  & 3.97 & 6.33 & 6.52 \\ 
        TimeSformer &  & - & - & - & - &  & 3.66 & 5.98 & 6.07 \\
        S3D & & - & - & - & - & & 3.57 & 5.91 & 5.49 \\
        \bottomrule
        \end{tabular}
    \end{minipage}
   \caption{(Top) Visual explanations revealing difficulty in player identification. Left: Although our detector captures the ball and player correctly, the face, jersey and size of the key player are barely recognizable due to blur. Middle: The detected player area is crowded and the ball handler is occluded by defenders. Right: A case where the ball is missing; thus, the model cannot find decisive information on the key player. (Bottom) Player identification results in percentage ($\%$) with our full approach and ablations on choice of features.}
\label{tab:identity_reco}
\end{figure}

We adopt the same training and evaluation strategy as in action recognition to measure the performance of our model on player identification, due to these tasks having the same format, i.e., a sequence of player names involved in the depicted action; Fig.~\ref{tab:identity_reco} has results. Resembling observations in the previous subsection, we find the quality of identified player names increases as we add more features and our full approach (top row) once again is the best performer. It also is seen that the results on all metrics are much worse than those of action recognition, cf., Table~\ref{tab:action_reco}. To explore this discrepancy, we study some failure cases in the images along the top of Fig.~\ref{tab:identity_reco}. It is seen that failure can be mostly attributed to blur, occlusion from unrelated regions and otherwise missing decisive information. 
\section{Conclusion}
In this work, we create a large-scale sports video dataset (NSVA) supporting multiple tasks: video captioning, action recognition and player identification. We propose a unified model to tackle all tasks and outperform the state of the art by a large margin on the video captioning task. 
The creation of NSVA only relies on webly data and needs no extra annotation. We believe NSVA can fill the opening for a benchmark in fine-grained sports video captioning, and potentially stimulate the application of automatic score keeping. 

The bottleneck of our model is player identification, which we deem the most challenging task in NSVA. To this end, a better algorithm is needed, e.g., opportunistic player recognition when visibility allows, with subsequent tracking for fuller inference of basketball activities. 
%There are two additional directions 
%we also plan to explore. 
There also are two additional directions we will explore: (1) We will investigate more advanced video feature representations (e.g., Video Swin transformer~\cite{video-swin}) on NSVA and compare to TimeSformer. (2) Prefix Multi-task learning~\cite{prefix-learning} has been proposed to learn several tasks in one model. Ideally, a model can benefit from learning to solve all tasks and gain extra performance boost on each task. We will investigate NSVA in the Prefix Multi-task learning setting with our task head. 
%The bottleneck of our model lies in its performance on the player recognition task, which we deem the most challenging task in NSVA

\noindent\textbf{Acknowlegement.} The authors thank Professor Adriana Kovashka for meaningful discussions in the early stages and Professor Hui Jiang for proofreading and valuable feedback. This research was supported in part by a NSERC grant to Richard P. Wildes and a CFREF VISTA Graduate Scholarship to He Zhao.
\clearpage
% ---- Bibliography ----
%
% BibTeX users should specify bibliography style 'splncs04'.
% References will then be sorted and formatted in the correct style.
%
\bibliographystyle{splncs04}
% \bibliography{eccv2022submission}
% \input{eccv2022submission.bbl}
\bibliography{egbib}
\end{document}